\documentclass[conference,letterpaper,10pt]{IEEEtran}
\usepackage[T1]{fontenc}
\usepackage{CJKutf8}
\makeatletter
\let\AstroBenchCJKaddcmap\CJK@addcmap
\def\CJK@addcmap#1{\AstroBenchCJKaddcmap{#1}\pdfnobuiltintounicode\font@name}
\makeatother
\usepackage{amsmath,amssymb,booktabs,tabularx,array,graphicx,cite,url}
\usepackage[hidelinks]{hyperref}
\newcolumntype{Y}{>{\raggedright\arraybackslash}X}
\newcommand{\pifive}{$\pi_{0.5}$}

\hypersetup{pdftitle={AstroBench: Benchmarking Free-Floating Capture and Space Robotic Manipulation},pdfauthor={Zizhe Wang, Jia Deng, Xin Lu, Zehao Jia, Yi Huang, Yao Zhu}}
\title{AstroBench: Benchmarking Free-Floating Capture\\and Space Robotic Manipulation}
\author{%
\IEEEauthorblockN{Zizhe Wang\textsuperscript{1}, Jia Deng\textsuperscript{1},
Xin Lu\textsuperscript{1}, Zehao Jia\textsuperscript{1}, Yi Huang\textsuperscript{2}, Yao Zhu\textsuperscript{2}}
\IEEEauthorblockA{\textsuperscript{1}Astroup Lab, Beijing, China\\
\textsuperscript{2}Harbin Institute of Technology, Beijing Research Institute}}
\begin{document}
\raggedbottom
\maketitle
\pagestyle{plain}
\thispagestyle{plain}
\begingroup
\renewcommand{\thefootnote}{}
\footnotetext{Astroup Lab: \begin{CJK*}{UTF8}{gbsn}向上跃升（北京）科技有限公司\end{CJK*}.\newline Homepage: \url{https://astroup.com/}.}
\endgroup

\begin{abstract}
Capturing a loose object inside a spacecraft requires a robot to approach a target that can translate and rotate, and whose motion changes when contact is made. These conditions motivate evaluation beyond manipulation of supported objects. We introduce AstroBench, an open benchmark for free-floating capture and space robotic manipulation with a fixed-base arm. The task suite combines four initial-motion tracks with sample-storage operations based on MELFI hardware and cap and captive-screw servicing based on the Robotic Refueling Mission. The two workcells provide nine task definitions and 29 task--target combinations. A metadata-filtered Objaverse inventory supplies 8,414 object instances for asset expansion, alongside five cabin visual contexts and versioned task instructions. We report two acquisition-focused experiments using existing checkpoints. A frozen $\pi_{0.5}$ DROID joint-position policy captures 2 of 200 static targets and 4 of 200 moving targets across twenty object identities. A cube-specific ACT policy captures 8 of 10 static targets and 6 of 10 moving targets in a separate experiment. Each experiment retains its recorded observation, action and capture configuration. The results document difficult direct transfer for the tested generalist checkpoint and successful task-specific capture behavior, while identifying concrete settings for studying motion-aware perception and contact-sensitive control. Code, task definitions and editable workcell assets are available at \url{https://github.com/AstroUpLab/AstroBench}.
\end{abstract}
\begin{IEEEkeywords}
Space robotics, free-floating capture, robot manipulation, benchmarks, vision-language-action models, imitation learning.
\end{IEEEkeywords}

\section{Introduction}
Loose-object recovery, sample handling and equipment servicing are useful capabilities for robots operating inside spacecraft. A robot may need to intercept a drifting tool, transfer a sample card into a storage box, or engage a service interface with a specialized tool. Although these tasks share components with terrestrial manipulation, their physical setup changes the role of support, approach direction and contact. A free object has no table to resist an incidental push. A storage or service interface instead provides an explicit fixture, whose geometry determines where the robot can move and how the operation should be completed.

Free-floating capture offers a particularly direct way to study this difference. Even a target initialized at rest can move after the fingers touch it. A translating or rotating target adds an interception problem: the robot must align the gripper with a useful grasp configuration before that configuration moves away. The visual observation, the action prediction and the resulting contact therefore correspond to different moments of a changing scene. Capture performance depends jointly on recognizing the object, approaching it, and coordinating contact with gripper closure.

Existing robot benchmarks have established effective ways to organize language-conditioned tasks, test transfer and vary environments~\cite{libero,calvin,robocasa}. Space Robotics Bench further develops robot-learning infrastructure for mission-relevant space environments~\cite{srb}. We build on this approach by focusing on free-floating object acquisition and a small set of hardware-grounded space operations. The benchmark separates the target object, initial motion, visual context and operation goal, so that each can be recorded and varied explicitly.

\begin{figure*}[t]
\centering
\includegraphics[width=\textwidth]{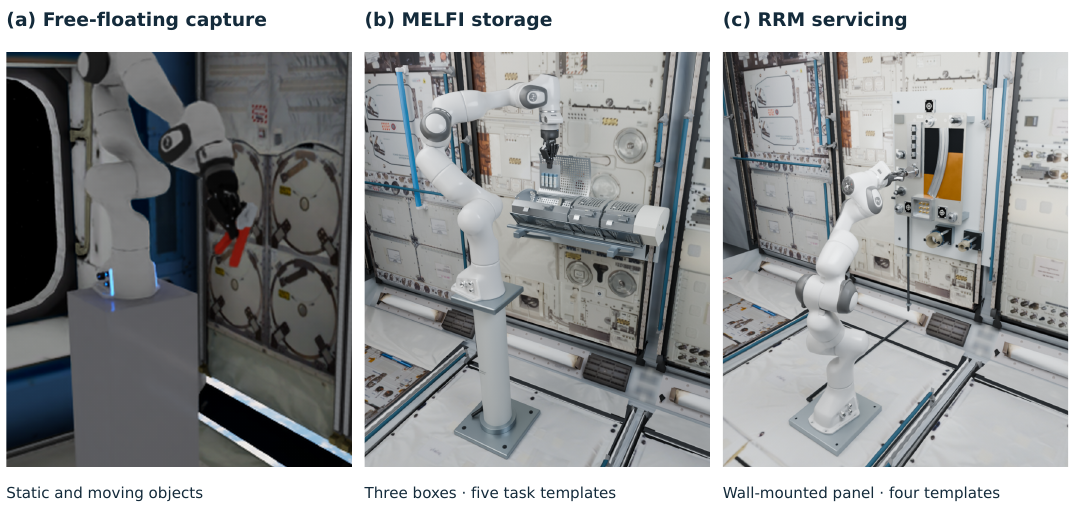}
\caption{AstroBench combines free-floating capture, MELFI sample-storage operations and RRM cap and fastener servicing. Panel (a) is a recorded \pifive{} rollout; panels (b) and (c) show the current workcell geometry. The capture target is unsupported; the storage and service hardware is secured to its worksite. The RRM panel is wall-mounted, with a floor-mounted arm centered in front of it.}
\label{fig:overview}
\end{figure*}

AstroBench contains three task groups (Fig.~\ref{fig:overview}). Free-floating capture uses static, drifting, tumbling and faster-tumbling initial conditions. A MELFI-inspired storage workcell supports card transfer, insertion and extraction, together with lid opening and closing. An RRM-inspired service workcell supports removal and stowage of FDV1 and SMA caps and loosening or tightening of captive screws. The hardware workcells are authored from NASA and ESA references, with identifiable moving parts, target frames and tool interfaces. The arm has a fixed base throughout; the free-floating body in the capture tasks is the target object.

This paper evaluates capture as an acquisition event: an episode succeeds when the target first satisfies the recorded capture detector. We examine two existing policy configurations. The first is a frozen \pifive{} DROID joint-position checkpoint evaluated on twenty objects, each under static and dynamic conditions. The second is ACT trained on a local cube task, evaluated at ten positions under both conditions. The experiments document what each tested system can capture and provide concrete traces for further analysis.

Our contributions are threefold. First, we organize a space manipulation suite that combines controlled initial object motion with storage and servicing tasks derived from real hardware references. Second, we release an extensible resource comprising an 8,414-instance object inventory, five cabin visual contexts, editable workcell assets and explicit language-conditioned task specifications. Third, we report acquisition results and reproducible execution settings for a generalist-policy transfer experiment and a task-specific imitation-learning experiment. Together, these resources support study of object motion, contact formation and precise operation goals in a common task framework.

\section{Related Work}
\textbf{Manipulation benchmarks.} LIBERO organizes robot-learning problems around transfer of task knowledge~\cite{libero}. CALVIN evaluates language-conditioned manipulation and long-horizon composition~\cite{calvin}. RoboCasa expands the diversity of everyday tasks and environments~\cite{robocasa}. These benchmarks demonstrate the usefulness of explicit task definitions, controlled variation and executable evaluation. Table~\ref{tab:related} summarizes their research emphasis. AstroBench uses those principles to structure free-floating capture and hardware-specific operations. Its object-motion tracks and fixture geometry make the physical conditions of each task explicit.

\textbf{Space robot learning.} Space Robotics Bench provides modular environments, procedural generation and robot-learning studies across space scenarios~\cite{srb}. AstroBench concentrates on the capture of unsupported objects and manipulation at specified storage and service interfaces. We use MELFI documentation to inform sample-storage geometry~\cite{melfi_esa,melfi_nasa}, and NASA's RRM material to inform cap, fastener and tool relationships~\cite{rrm_nasa,rrm_2013}. These sources ground the choice of tasks and visible hardware. The robot placement and evaluation parameters are engineering choices for the benchmark.

\textbf{Visual policies.} ACT predicts action chunks from images and robot state~\cite{act}. The \pifive{} model investigates broad robotic generalization using vision, language and action~\cite{pi05}, and OpenPI provides model implementations and deployment configurations~\cite{openpi}. We use these two approaches to obtain complementary capture measurements: direct transfer of an existing DROID checkpoint, and execution of a policy trained for the cube task. The present contribution is the benchmark and its empirical characterization; no new policy architecture is introduced.

\textbf{Object and environment assets.} Objaverse offers a large collection of annotated 3D objects~\cite{objaverse}. Its scale makes metadata screening practical, but an asset name does not determine whether its geometry is suitable for a particular gripper or contact model. AstroBench therefore preserves object identities, selection reasons and source attribution separately from runtime asset configuration. NASA Astrobee ISS assets supply the cabin visual contexts~\cite{astrobee,astrobee_assets}. Task bodies and robot mounting are described independently of those surroundings.

\begin{table}[t]
\centering\footnotesize\setlength{\tabcolsep}{3pt}
\caption{Research emphasis of representative related benchmarks. This comparison describes the questions addressed rather than ranking platform performance.}
\label{tab:related}
\begin{tabularx}{\columnwidth}{lY}\toprule
Benchmark & Emphasis relevant to this work\\\midrule
LIBERO & Transfer across structured manipulation suites\\
CALVIN & Language-conditioned, long-horizon composition\\
RoboCasa & Diverse everyday tasks and visual environments\\
Space Robotics Bench & Mission-relevant robot learning in space environments\\
AstroBench & Free-floating capture and reference-based storage/service interfaces\\\bottomrule
\end{tabularx}
\end{table}

\section{Benchmark Design}
\subsection{Physical Setting}
The reference environment uses a fixed robot base and zero gravity. The recorded capture experiments use a Panda arm with a Robotiq gripper. Target objects are rigid bodies with no supporting surface. An initial pose and twist are applied at reset; subsequent target motion follows the environment's rigid-body dynamics and robot contact. Thus, ``static'' refers to an initial condition rather than an attachment that prevents later motion.

The workcells use a second contact setting: storage boxes and service hardware are mounted to fixtures. The robot operates their movable parts or transfers an object relative to the fixture. Keeping these two settings explicit avoids treating every space task as a drifting-object problem. Capture emphasizes interception and contact formation; the workcells emphasize target selection, alignment, constrained motion and operation-specific endpoints.

\subsection{Factorized Task Specification}
A task instance is represented by
\begin{equation}
 e=(\tau,o,z,x_0,c,\ell,\nu),
\end{equation}
where $\tau$ is the task definition, $o$ identifies the object or hardware asset, $z$ is the selected target, $x_0$ is the initial state, $c$ is the visual context, $\ell$ is the resolved instruction, and $\nu$ identifies the execution and scoring configuration. For capture, $z$ is the object; for a workcell, it can be one box, a particular cap or a captive screw.

The specification binds task identity to physical and language identity. Changing the selected cap changes the target frame and instruction. Changing an opening task into a closing task changes the initial mechanism state as well as the goal. Repeating an episode changes its run identity without implying a different geometry or randomized initial pose. These distinctions are important when interpreting sample counts and designing generalization studies.

\subsection{Hardware Reference and Robot Access}
The workcells follow visible components, dimensions and procedures documented by NASA and ESA. Models separate lids, latches, caps, fasteners, adapters and mounting structures so that their task roles remain accessible to an environment adapter. Collision descriptions preserve openings and tool interfaces; a single solid approximation would remove the very geometry that makes insertion or engagement meaningful.

The hardware layout is designed together with robot access. MELFI boxes run parallel to the cabin wall, with an arm mounted on a floor-supported column in front of the openings. The RRM panel is mounted vertically and the arm is placed directly in front on a thin floor plate. This RRM installation is a cabin task arrangement of servicing hardware originally developed for an exterior mission. The modeling reference and the authored experimental installation are recorded separately.

\section{Task Suite and Assets}
\subsection{Free-Floating Capture}
The capture task asks the robot to acquire a named object. The released instruction template is \emph{``Capture the \{object\_name\} and hold it securely.''} The instruction is kept unchanged across motion tracks. The endpoint analyzed in this paper is the acquisition event defined in Section~\ref{sec:protocol}; the archived instructions are preserved verbatim.

Table~\ref{tab:motion} describes the four initial-motion tracks in the task catalog. Motion is applied once at reset. The seed and object identity determine the signs of translation along world $x$ and rotation about world $z$. These tracks control two motion components while retaining the same object and task wording. The reported historical experiments use their recorded static/dynamic settings, which are specified separately in Section~\ref{sec:experiments}.

\begin{table}[t]
\centering\footnotesize
\caption{Initial-motion tracks in the released capture catalog. Magnitudes are shown; registered seeds determine the signed directions.}
\label{tab:motion}
\begin{tabular}{lrr}\toprule
Track & Linear speed (m/s) & Angular speed (rad/s)\\\midrule
Static & 0 & 0\\
Drift & 0.02 & 0\\
Tumble & 0.02 & 0.5\\
Fast tumble & 0.05 & 1.0\\\bottomrule
\end{tabular}
\end{table}

The compact collection contains twenty authored object identities: geometric parts, containers, tools and portable devices. Examples include a sample tube, pliers, a flashlight, a ring and a spool. Their distinct geometries provide different approach surfaces, grasp widths and contact configurations. The twenty-object catalog crossed with four tracks yields eighty object--motion cases per repetition. Object expansion is handled through the same asset and task interface.

\subsection{MELFI Sample-Storage Workcell}
MELFI is a space-station sample-storage facility. ESA's description documents a tray arrangement containing two quarter boxes and one half box, and NASA's interface material describes boxes, lids and handling features~\cite{melfi_esa,melfi_nasa}. AstroBench reconstructs this arrangement as two narrow boxes beside a wider box, with separate lids, locks, sample-card geometry and internal details. The sample card and its interaction parameters are benchmark engineering choices informed by those references.

Five tasks cover different stages of sample handling: transfer a card to the selected opening, insert it, extract it, open the lid and close the lid. Each task addresses one of three boxes, yielding fifteen task--target combinations. The insertion task starts with the card held outside the box; extraction starts with it inside. Transfer starts with an open gripper and a card outside the box. This explicit initialization lets each task be evaluated independently.

Geometric goals refer to the selected box and the entire card mesh. Moving the card center inside the box is insufficient if part of the card remains outside. Opening and closing use lid and latch states, while the extraction goal includes clearance from the opening. Table~\ref{tab:tasks} summarizes the task endpoints; exact language templates and targets are given in the appendix.

\subsection{RRM Cap and Fastener Workcell}
NASA's Robotic Refueling Mission demonstrated servicing operations involving specialized tools and satellite-like interfaces~\cite{rrm_nasa,rrm_2013}. AstroBench's TB1-inspired panel includes an FDV1 tertiary cap, a $3\times3$ SMA cap array and two captive-screw positions. The FDV1 task uses a Multifunction Tool with a Tertiary Cap Adapter. SMA operations use a Safety Cap Tool with an SMA adapter; captive screws use the corresponding torque-set adapter.

Cap removal includes tool engagement, unscrewing, axial withdrawal and docking of the cap--adapter assembly in its receiver before driver withdrawal. The selected SMA cap is identified by its physical row and column, providing a concrete language-grounding problem among visually related targets. In the current vertical layout, rows increase upward and columns increase from left to right when facing the panel. Captive-screw loosening and tightening are separate tasks with distinct initial states; the screw remains part of the panel mechanism.

The four task definitions expand to fourteen combinations: one FDV1 target, nine SMA targets, and two screws for each of loosening and tightening. Together with MELFI, the workcells contain nine definitions and twenty-nine combinations. These counts refer to task and target identities, rather than twenty-nine separately modeled environments.

\begin{table*}[t]
\centering\footnotesize\setlength{\tabcolsep}{4pt}
\caption{Workcell tasks and their operation goals. The target count expands each task definition into concrete language-conditioned cases.}
\label{tab:tasks}
\begin{tabularx}{\textwidth}{llrYY}\toprule
Workcell & Task & Targets & Initial condition & Goal evidence\\\midrule
MELFI & Card transfer & 3 & Card outside; gripper open & Card acquired and aligned with the selected opening\\
MELFI & Card insertion & 3 & Card held outside the box & Whole card inside; gripper released; storage restraint engaged\\
MELFI & Card extraction & 3 & Card inside the box & Card acquired and clear of the opening\\
MELFI & Lid opening & 3 & Lid closed and latched & Latch released and lid at the open angle\\
MELFI & Lid closing & 3 & Lid open & Lid closed and latched; tool clear\\\midrule
RRM & FDV1 removal and stowage & 1 & Cap seated and threaded & Cap removed with its adapter; assembly docked; driver withdrawn\\
RRM & Selected SMA removal and stowage & 9 & Selected cap seated and threaded & Requested cap removed and docked with its adapter\\
RRM & Captive-screw loosening & 2 & Screw tightened & Required rotation and axial travel; screw remains captive\\
RRM & Captive-screw tightening & 2 & Screw loosened & Screw seated; tool disconnected and clear\\\bottomrule
\end{tabularx}
\end{table*}

\subsection{Object Inventory and Cabin Contexts}
Objaverse is the primary source for the expansion inventory~\cite{objaverse}. A versioned category-and-name filter scans 798,759 metadata records and selects rigid tools, containers, portable electronics, mechanical parts and relevant supplies. Whole scenes, buildings, furniture, complete vehicles, organisms and other unsuitable entries are filtered out. The resulting inventory contains 8,414 object instances across 51 LVIS labels and additional unlabeled objects. Each selected record retains its source identifier, name, attribution, license metadata and selection reason.

This inventory is an asset-selection resource. Instantiating a scored object also requires scale, collision geometry, mass, inertia and gripper compatibility. Those properties belong to the runtime asset record and are bound to the episode. Keeping selection metadata and physical instantiation separate allows new assets to enter the same task protocol without treating a category label as a physical specification.

Five cabin contexts are registered: Kibo/JPM, the US Laboratory, the European Laboratory, Node 2, and a Node 3 worksite near the Cupola. They use NASA Astrobee visual assets~\cite{astrobee,astrobee_assets}. The context can be changed while preserving task bodies and initial conditions. This supports visual-context studies without coupling every background change to a new object or mechanism. The \pifive{} experiment reported here uses Kibo/JPM.

\section{Evaluation Protocol}\label{sec:protocol}
\subsection{Capture as the Primary Endpoint}
Let $C_k$ be the capture predicate evaluated from measured state at step $k$, and let $K_i$ be the last available evaluation step for attempt $i$. Its acquisition outcome is
\begin{equation}
 A_i=\mathbf{1}\left[\exists\,k\in\{0,\ldots,K_i\}: C_k=1\right].
\end{equation}
For $N$ registered attempts, the capture rate is
\begin{equation}
 \mathrm{CR}=\frac{1}{N}\sum_{i=1}^{N} A_i.
\end{equation}
The first qualifying event time is $t_i^{\mathrm{cap}}=\min\{t_k:C_k=1\}$. Counts and rates are reported by motion condition and object, with the aggregate denominator stated explicitly.

The detector uses actual target contact and gripper state, together with the geometry and relative motion required by its recorded revision. A close command or a visually overlapping projection alone is insufficient. A short temporal qualification is part of contact-event detection; its configured value is documented with each experiment. The acquisition fields are extracted from the original records without changing detector thresholds after inspecting outcomes. The reported capture rates are reconstructed from native acquisition events, independently of the public tool's default \texttt{strict\_success} summary field.

The two historical experiments use different detector revisions. We preserve those revisions and report each experiment independently. The released capture planner uses a twelve-second specification and four motion tracks; the archived \pifive{} and ACT executions use the horizons and two-condition settings listed in Table~\ref{tab:config}. This paper's empirical endpoint is acquisition, while the resource catalog describes the current task definitions.

\subsection{Measured Geometry and Task-Specific Goals}
Judging requires a consistent reference frame. If $p_o$ and $p_g$ are target and gripper reference points in world coordinates and $R_g$ maps gripper to world coordinates, the relative target position is
\begin{equation}
 r=R_g^\top(p_o-p_g).
\end{equation}
The corresponding relative linear velocity at the target reference point is
\begin{equation}
 v_{\mathrm{rel}}=R_g^\top\left[v_o-v_g-\omega_g\times(p_o-p_g)\right].
\end{equation}
The rotating-frame term matters when the wrist turns during approach. The environment adapter declares whether the target reference is a center of mass or a geometry center and samples the contact and state signals at the same step.

Workcell success is defined by the selected operation. Card placement uses signed containment and opening-clearance measurements for the complete mesh. Lid tasks use actual hinge and latch states. Cap operations use accumulated fastener angle, axial travel, adapter and receiver states, and driver clearance. Captive-screw tasks use rotation and travel while checking that the screw remains attached. A successful intermediate grasp therefore does not substitute for the goal of an operation task.

\subsection{Plans, Instructions and Attempt Accounting}
Episode plans include task, target, asset, initial state and language identities. The instruction is resolved before policy execution. A task specification hash binds those fields, and the recorded trace identifies the same episode. The public task registry can export all concrete instructions, including target-specific SMA prompts. Identical instruction text across motion tracks holds the language request fixed while initial motion changes.

Registered attempts remain in the accounting even when the action interface rejects a prediction. Such outcomes are separated from completed execution without acquisition. Measurement and infrastructure errors are recorded as distinct statuses so that missing evidence cannot become an apparent successful run. Repeated nominal states and randomized test distributions are also distinguished: the former estimate execution variation at specified states, whereas the latter address coverage of a distribution.

\section{Experiments and Analysis}\label{sec:experiments}
\subsection{Questions and Recorded Configurations}
Both capture studies were conducted in NVIDIA Isaac Sim, using Isaac Lab-based scene and control interfaces with the GPU PhysX backend. The experiments address two questions. First, how often does a frozen DROID \pifive{} configuration acquire the free-floating objects in the recorded twenty-object evaluation? Second, can a policy trained for the cube task acquire static and moving targets at the tested positions? We report the recorded static/dynamic settings rather than treating them as executions of every track in the current catalog.

\begin{table*}[t]
\centering\footnotesize\setlength{\tabcolsep}{4pt}
\caption{Recorded capture configurations. Each column defines its own experiment; object coverage, initial states, detectors and action scheduling differ.}
\label{tab:config}
\begin{tabularx}{\textwidth}{lYY}\toprule
Setting & Frozen \pifive{} DROID & Cube-specific ACT\\\midrule
Training use & Existing checkpoint; no adaptation in this evaluation & 400 local cube demonstrations; trained from random initialization; frozen at step 140,000\\
Objects & Twenty authored object identities & One Rubik's cube\\
Initial states & One nominal pose per object, repeated ten times per condition & Ten fixed development positions, each evaluated once per condition\\
Conditions & Static / translation plus rotation & Static / translation plus rotation\\
Moving-target initialization & 0.006 m/s along world $x$; 0.08 rad/s about world $z$ & 0.006 m/s along world $x$; 0.08 rad/s about world $z$\\
Physics / action rate & 120 Hz / 15 Hz & 120 Hz / 15 Hz\\
Action schedule & Fifteen-action chunks executed before replanning & Thirty-action chunks; inference every action step; temporal ensembling\\
Episode duration & 25 s & 20 s\\
Policy observations & External and wrist RGB, joint state, gripper state, instruction & Recorded RGB views and proprioception; task-specific policy\\
Action representation & Seven absolute joint targets and binary gripper command & Absolute joint targets and binary gripper command\\
Capture detector & Local target contacts and palm-relative geometry; 1 s qualification & Opposing target contacts and between-finger geometry; 15 qualifying samples\\
Number of attempts & 400 & 20\\\bottomrule
\end{tabularx}
\end{table*}

The \pifive{} configuration is \path{pi05_droid_jointpos_polaris}, with the recorded DROID joint-position weights. External and wrist images are padded and resized to $224\times224$. The adapter executes the first fifteen returned actions over one second before replanning. Joint-limit checks reject invalid target commands rather than silently altering them. The checkpoint is frozen throughout the experiment.

The ACT model was trained from random initialization on 400 local demonstrations selected from earlier \pifive{} cube rollouts, totaling 58,925 frames at 15 Hz. The evaluated checkpoint is step 140,000. It uses its original image preprocessing, a thirty-action prediction chunk and temporal ensembling with coefficient 0.01. Inference occurs at every action step. ACT is therefore a task-specific learned baseline, with a training and execution regime distinct from the \pifive{} transfer experiment.

\subsection{Frozen \texorpdfstring{\pifive{}}{pi0.5} Capture Results}
The evaluation contains twenty objects, two motion conditions and ten repetitions per object--condition pair. Table~\ref{tab:pi05} reports all four hundred attempts. The policy acquires two static targets and four dynamic targets, giving capture rates of 1\% and 2\%, respectively. The overall rate is 6/400, or 1.5\%.

\begin{table}[t]
\centering\footnotesize\setlength{\tabcolsep}{4pt}
\caption{\pifive{} acquisition results. Rejections are included in the attempt denominator and shown separately.}
\label{tab:pi05}
\begin{tabular}{lrrrr}\toprule
Condition & Attempts & Captures & Rate & Rejections\\\midrule
Static & 200 & 2 & 1.0\% & 0\\
Dynamic & 200 & 4 & 2.0\% & 2\\\midrule
Total & 400 & 6 & 1.5\% & 2\\\bottomrule
\end{tabular}
\end{table}

Static captures occur for the small hammer and sample tube, once each. All four dynamic captures occur for the pliers. First capture is recorded between 11.47 and 24.33 s across the six successful attempts. The two dynamic action rejections occur for a cylindrical part and a sealed bottle; neither is counted as acquisition. Of the 398 attempts that execute without an action-limit rejection, 392 do not reach the capture predicate.

These observations establish low acquisition performance for this checkpoint and deployment configuration on the tested object set. The concentration of captures on three objects motivates per-object reporting: a single aggregate would obscure the uneven outcome across geometries. The counts do not establish that dynamic targets are generally easier, despite the larger dynamic count in this run. The two conditions contain few successes, and repeated attempts reuse nominal initial states.

\subsection{Task-Specific ACT Capture Results}
ACT is evaluated on the cube at ten fixed positions under static and moving conditions. It acquires eight static targets and six moving targets, for fourteen acquisitions in twenty attempts (Table~\ref{tab:act}). The same checkpoint and inference setup are used across the two conditions. This experiment demonstrates capture behavior learned for a specific target and reports the observed reduction in acquisitions when initial translation and rotation are introduced together.

\begin{table}[t]
\centering\footnotesize
\caption{Cube-specific ACT acquisition results in the 20 s experiment.}
\label{tab:act}
\begin{tabular}{lrrr}\toprule
Condition & Attempts & Captures & Rate\\\midrule
Static & 10 & 8 & 80\%\\
Moving & 10 & 6 & 60\%\\\midrule
Total & 20 & 14 & 70\%\\\bottomrule
\end{tabular}
\end{table}

\begin{figure*}[t]
\centering
\includegraphics[width=\textwidth]{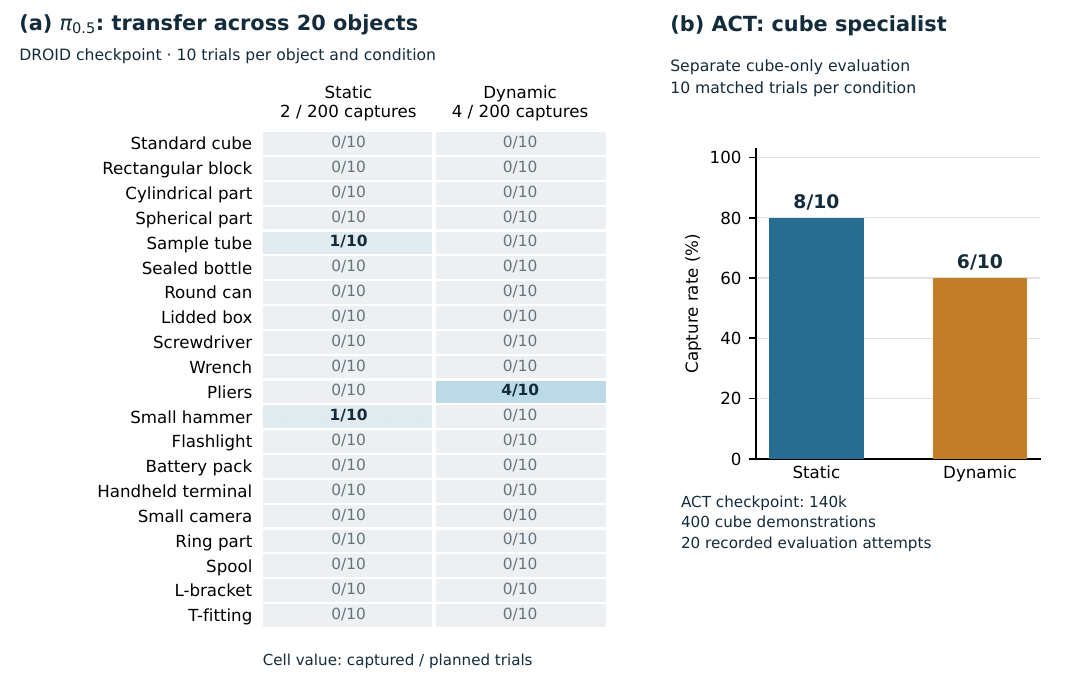}
\caption{Acquisition measurements from the two recorded experiments. The \pifive{} panel resolves the twenty-object evaluation by object and motion condition; each cell has ten attempts. ACT is shown in a separate panel for its cube-specific experiment. Table~\ref{tab:config} specifies the settings of each panel.}
\label{fig:results}
\end{figure*}

\begin{figure*}[t]
\centering
\includegraphics[width=\textwidth]{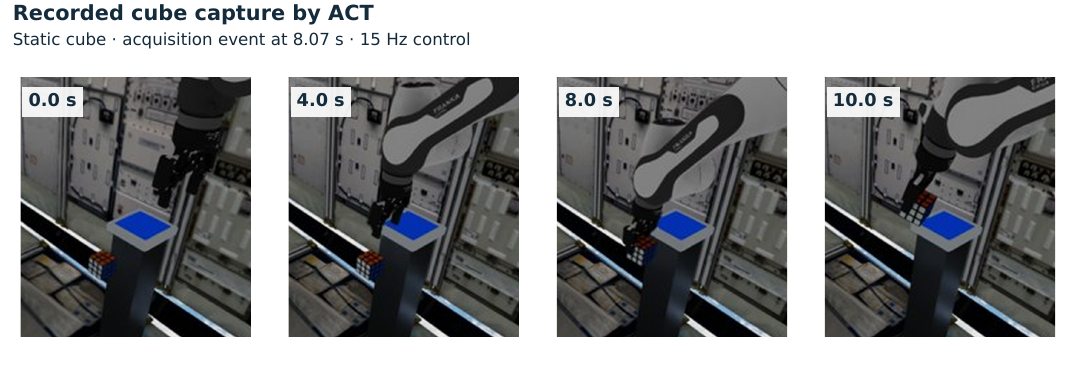}
\caption{Recorded ACT capture of an initially stationary cube, shown at 0, 4, 8 and 10 s in one of the twenty evaluation attempts. The measured capture predicate first qualifies at 8.07 s. Frames illustrate approach and acquisition; the event time is obtained from the synchronized state trace rather than inferred from these sparsely sampled images.}
\label{fig:sequence}
\end{figure*}

The ten positions are development positions, not a held-out geometry benchmark. The result therefore characterizes the deployed cube policy at those states. It is not used to estimate capture on the 8,414-instance inventory. Likewise, the difference between the ACT and \pifive{} percentages is not an architecture comparison: task coverage, training data, detector revision, horizon and action scheduling differ. Reporting the two experiments separately preserves their respective questions. Figure~\ref{fig:sequence} illustrates one recorded ACT approach and acquisition; aggregate outcomes are shown in Fig.~\ref{fig:results}.

\subsection{What the Acquisition Evidence Reveals}
The \pifive{} records distinguish failure to acquire during a completed rollout from rejection of an action command. Only two attempts terminate at the joint-limit interface, so this event alone does not account for the low capture count. The remaining non-acquisitions require trajectory-level analysis. Their summary flags do not identify whether perception, approach alignment, closure timing or contact-induced motion is the dominant cause.

The action schedules suggest a concrete mechanism to investigate. The \pifive{} deployment executes one second of predicted actions before receiving a new observation, whereas ACT recomputes a temporally ensembled command at each control step. During that interval, a moving target changes pose and a contact can redirect it. This is a testable hypothesis about the interaction between feedback rate and capture, rather than a causal conclusion from the two model totals. A within-policy scheduling experiment would isolate it more directly.

The physical setup offers a second useful distinction. A static free object can still be displaced by the approaching gripper; its initial rest condition does not supply a support reaction. A matched support-condition experiment would help separate that effect from changes in visual appearance, object geometry and embodiment. In the current results, these factors remain part of the transfer setting. The task specification records them so that future experiments can vary them one at a time.

\section{Discussion and Open Resources}
\subsection{From Capture to Hardware Operations}
The capture experiments provide empirical measurements for one part of the suite. MELFI and RRM extend the task vocabulary toward sample handling and servicing, where target selection and interface geometry are explicit. Their current evidence consists of editable models, reset definitions, geometric access checks and recorded mechanism demonstrations. These artifacts support inspection of the task design; the policy results reported in this paper concern capture.

The current RRM wall layout has a sampled CAD and inverse-kinematics review covering the ready pose, transformed mechanism paths, and approaches to all twelve physical operation targets. The model records contain 335 checked poses. MELFI has a card-to-opening geometry check. These checks address reachability and modeled interference at sampled poses. They are kept separate from measured policy episodes and from the original tabletop RRM mechanism demonstrations.

\subsection{Reproducibility and Research Use}
The task catalog described here corresponds to release 0.4.0 (repository commit \texttt{e4619eb6}). The public release contains task and language registries, object-selection metadata, deterministic plan generation, measurement-based scoring tools, and editable Blender and GLB workcell assets. Semantic parts, collision descriptions and robot attachment frames connect a task definition to an environment implementation. Asset hashes and protocol identities record what was evaluated, while explicit observation and action settings make policy deployment inspectable.

The suite supports several focused studies: interception under separately varied translation and rotation; capture across object geometry; target selection among repeated service interfaces; and visual transfer across cabin contexts. For comparisons across models, a shared object set, state distribution, detector, horizon and action-interface specification should be registered. For geometry transfer, related mesh variants should be assigned to the same split. Neither repeated nominal states nor a large asset inventory alone establishes generalization.

The present setting isolates fixed-base manipulation and uses authored contact and mechanism models. Reference photographs and engineering documents inform hardware appearance and task sequence, while collision approximations, friction and mechanism parameters remain part of the experimental specification. This division makes the resource useful for controlled robot-learning experiments without assigning flight-hardware fidelity to unmeasured properties.

\section{Conclusion}
AstroBench organizes free-floating capture and space robotic manipulation around explicit objects, motion conditions and operation goals. It combines an expandable 8,414-instance inventory with MELFI sample-storage and RRM servicing workcells, task-specific language instructions and reproducible evaluation records. Acquisition-focused experiments report 6/400 captures for a frozen \pifive{} DROID configuration and 14/20 for a separately evaluated cube-specific ACT policy. These measurements provide concrete baselines for their recorded settings and motivate controlled studies of motion perception, feedback scheduling and contact formation in space-oriented manipulation.

\appendices
\section{Experiment Configuration and Per-Object Results}
\label{app:experiments}

\subsection{Capture Endpoint and Evidence}
An episode counts as a capture when its recorded native grasp predicate is satisfied at least once within its execution budget. We recomputed this event and its first occurrence from the measured traces, and checked them against the original episode receipts: 149,870 samples for the 400 $\pi_{0.5}$ attempts and 5,792 samples for the 20 ACT attempts. All receipts agree with their trace-derived capture events. The execution logs identify an NVIDIA GeForce RTX 4090 with driver 580.95.05 for both studies. The two experiments used different historical observation, execution and grasp-predicate configurations, specified below, and are reported separately.

Both native predicates require measured gripper closure above 0.05 rad, target-specific contact on both fingers above 0.2 N, opposing forces, contact geometry consistent with the object and fingers, and fresh contact measurements. Their capture-confirmation windows constrain relative displacement to 10 mm and relative rotation to 0.2 rad. The $\pi_{0.5}$ implementation uses local oriented bounding boxes with a 5 mm contact margin and 16 samples at 15 Hz, spanning 1 s. It does not impose an instantaneous relative-velocity threshold. ACT uses its original geometry checks and 15 consecutive control samples; its additional speed gates are 0.05 m/s and 1 rad/s. These are the predicates used to obtain the reported counts.

\subsection{\texorpdfstring{$\pi_{0.5}$}{pi0.5} Object Study}
We use OpenPI configuration \path{pi05_droid_jointpos_polaris}, with the recorded \path{pi05_droid_jointpos} checkpoint and DROID normalization statistics, without fine-tuning. The checkpoint identity is recorded by the 26-file storage-generation manifest. An external and a wrist RGB camera render at $640\times480$; full frames are resized with padding to $224\times224$. The state input contains seven measured arm joint positions and normalized gripper position. Actions specify seven absolute joint positions in radians and a binary gripper target. A 15-action prefix is executed at 15 Hz before replanning. Physics runs at 120 Hz and each attempt has a 25 s budget.

The robot is a fixed-base Franka Panda with a Robotiq 2F-85 gripper in the JPM cabin. Gravity is zero. Each of the 20 objects is tested ten times at an initially stationary state and ten times with initial velocity $\mathbf{v}=(0.006,0,0)$ m/s and $\omega=(0,0,0.08)$ rad/s. Each object uses one authored pose, shared by its two motion conditions; the ten repetitions reuse that pose. All recorded initial pose components are identical within each object--motion group. The policy server's random stream continues across requests, so the repetitions are not independently seeded pose samples.

Table~\ref{tab:app_pi05_objects} includes every registered grasp attempt. A transport failure before any action was issued was preserved and resumed under the same episode identity; it is not an additional manipulation attempt. Two dynamic attempts encountered predicted joint-limit violations and remain in the denominator. Capture occurred in 2/200 static attempts and 4/200 dynamic attempts, or 6/400 overall. The static captures were a small hammer at 17.47 s and a sample tube at 17.20 s. All four dynamic captures involved pliers, at 22.20, 15.87, 11.47 and 24.33 s in their respective repetitions.

\begin{table}[t]
\centering\footnotesize
\setlength{\tabcolsep}{4pt}
\caption{$\pi_{0.5}$ per-object capture counts. Each motion condition has ten attempts per object. Rej. counts predicted joint-limit rejections, included in the attempt denominators.}
\label{tab:app_pi05_objects}
\begin{tabular}{lrrrr}
\toprule
& \multicolumn{2}{c}{Captures / attempts} & \multicolumn{2}{c}{Rej.}\\
\cmidrule(lr){2-3}\cmidrule(lr){4-5}
Object & Static & Dynamic & S & D\\
\midrule
Cube & 0/10 & 0/10 & 0 & 0 \\
Rectangular block & 0/10 & 0/10 & 0 & 0 \\
Cylinder & 0/10 & 0/10 & 0 & 1 \\
Sphere & 0/10 & 0/10 & 0 & 0 \\
Sample tube & 1/10 & 0/10 & 0 & 0 \\
Sealed bottle & 0/10 & 0/10 & 0 & 1 \\
Round can & 0/10 & 0/10 & 0 & 0 \\
Lidded box & 0/10 & 0/10 & 0 & 0 \\
Screwdriver & 0/10 & 0/10 & 0 & 0 \\
Wrench & 0/10 & 0/10 & 0 & 0 \\
Pliers & 0/10 & 4/10 & 0 & 0 \\
Small hammer & 1/10 & 0/10 & 0 & 0 \\
Flashlight & 0/10 & 0/10 & 0 & 0 \\
Battery pack & 0/10 & 0/10 & 0 & 0 \\
Handheld terminal & 0/10 & 0/10 & 0 & 0 \\
Small camera & 0/10 & 0/10 & 0 & 0 \\
Ring & 0/10 & 0/10 & 0 & 0 \\
Spool & 0/10 & 0/10 & 0 & 0 \\
L-bracket & 0/10 & 0/10 & 0 & 0 \\
T-fitting & 0/10 & 0/10 & 0 & 0 \\
\midrule
Total & 2/200 & 4/200 & 0 & 2\\
\bottomrule
\end{tabular}
\end{table}

\subsection{Task-Trained ACT Study}
ACT was trained from random initialization on 400 local demonstrations of one task, ``Pick up the Rubik's cube and hold it securely.'' The demonstrations were selected from earlier local $\pi_{0.5}$ rollouts. The dataset contains 58,925 frames at 15 Hz; all 400 demonstrations were used for training. Training continued from 50,000 updates toward 150,000, and the evaluation freezes checkpoint 140,000 (SHA-256 prefix \texttt{b8477731b38d}). It is a task-trained ACT policy, with neither a pretrained policy nor pretrained visual-backbone initialization.

The policy receives seven joint positions, gripper state and two RGB views. Each $1280\times720$ view is center-cropped to $720\times720$ and bilinearly resized to $256\times256$, followed by the checkpoint's normalization. ACT predicts 30 actions, performs inference at every 15 Hz control step and uses temporal ensembling with coefficient 0.01. The robot and physics rate are as above; the execution budget is 20 s.

The study uses one cube and ten previously used development positions, each under static and dynamic initial motion. The dynamic velocities are the same as in the object study. The authored XY ranges are $x\in[0.55017,0.58886]$ m and $y\in[-0.09425,-0.06022]$ m. The ten paired reset seeds are 32,000,000 through 32,000,009; each seed is shared by its static and moving case. The original 20 episode receipts give 8/10 static and 6/10 dynamic captures (14/20 overall). No additional training was performed for these evaluations. Differences in task exposure, initial states, images, action execution and capture gates prevent interpreting the two studies as a controlled model ranking.

% Canonical prompts generated from the public language registry, not paraphrased.
% Source: release/astrobench-front-public/astrobench/data/task_prompts.json
% Registry SHA-256: 6d125fd31a1f2e51041e9598ca26bcc33914d4910db781e4839235433d34d932
\section{Hardware tasks and canonical language instructions}
\label{app:hardware_tasks}

\subsection{Hardware references and benchmark adaptations}
\paragraph{MELFI sample handling}
The tray arrangement follows the two quarter-size and one half-size box modules
shown in Fig.~6 of the ESA MELFI description. The same source describes a
$100\times110$\,mm aluminium vial card, elastic loops for vials, and elastic
bands that secure cards inside the box modules~\cite{melfi_esa}.
NASA SSP~57000 provides the box envelopes and distinguishes the quarter- and
half-size modules~\cite{melfi_nasa}. These references motivate the mixed tray,
card-handling sequence, and need for retention after release. AstroBench's
part meshes, latch mechanism, selected vial arrangement, robot grasp interfaces,
elastic parameters, and scoring tolerances are explicit engineering choices.
The model represents this sample-handling interface, not the freezer's thermal
operation or a validated replica of its contact dynamics.

\paragraph{RRM tool-mediated servicing}
NASA's RRM descriptions distinguish cap removal with a Multifunction Tool and
adapters from the Safety Cap Tool's cap, screw, and stowage operations
\cite{rrm_nasa,rrm_thermal}. The 2013 operations report identifies the SMA
adapter and Torque Set Adapter task sequence, including caged-screw
unfastening~\cite{rrm_2013}. AstroBench uses these interfaces to define FDV1
cap removal, addressed SMA-cap removal, and two directions of captive-screw
operation. Each cap task ends with the cap retained in the detached adapter,
the adapter secured in its holder, and the driver withdrawn. Merely moving a
cap off its thread is therefore insufficient.

The released RRM workcell is an \emph{author-designed intravehicular installation}
of this servicing hardware: the panel is vertical, its backplate is fixed to a
wall through four stand-offs, and the arm is mounted directly on a thin floor
plate in front of the panel. The original RRM was an external ISS experiment
operated with Dextre~\cite{rrm_thermal}; the cabin installation does not claim to
reproduce that flight mounting. Tool geometry, adapter couplings, thread
parameters, and the fixed-base layout are documented benchmark choices.

\subsection{Task--target expansion}
The registry \texttt{astrobench.language/1} contains nine hardware-task templates.
Table~\ref{tab:canonical_workcell_prompts} reproduces their English text exactly.
Substitution of the target names in Table~\ref{tab:canonical_workcell_targets}
gives all 29 instructions: $5\times3=15$ MELFI combinations and
$1+9+2\times2=14$ RRM combinations. IDs refer to persistent asset parts rather
than camera-relative image positions. In the wall-mounted RRM configuration,
SMA rows increase from bottom to top along panel-local $+Y$ (world $+Z$), and
columns increase from left to right along panel-local $+X$ when facing the panel.

\begin{table*}[t]
\centering
\caption{Canonical workcell instruction templates. Resolve each target placeholder
using the exact names in Table~\ref{tab:canonical_workcell_targets}. The labels M1--R4 are local table
abbreviations; task IDs are the machine-readable registry keys.}
\label{tab:canonical_workcell_prompts}
\footnotesize
\setlength{\tabcolsep}{4pt}
\renewcommand{\arraystretch}{1.08}
\begin{tabular}{@{}p{0.04\textwidth}p{0.27\textwidth}p{0.64\textwidth}@{}}
\toprule
Label & Task ID & Exact English prompt template \\
\midrule
M1 & \texttt{card\_transfer\_to\_mouth} & Move the card to the opening of the \{target\_name\} and hold it aligned. \\
M2 & \texttt{card\_insert} & Insert the card fully into the \{target\_name\}, release it, and leave it retained inside. \\
M3 & \texttt{card\_extract} & Remove the card from the \{target\_name\} and hold it clear of the opening. \\
M4 & \texttt{lid\_open} & Open the lid of the \{target\_name\} and leave it open. \\
M5 & \texttt{lid\_close} & Close and latch the lid of the \{target\_name\}, then withdraw the gripper. \\
R1 & \texttt{fdv1\_remove\_stow} & Unscrew and remove the \{target\_name\}, retain it in its adapter, dock the adapter in its stowage holder, and withdraw the driver. \\
R2 & \texttt{sma\_remove\_stow} & Unscrew and remove the \{target\_name\}, retain it in its adapter, dock the adapter in its stowage holder, and withdraw the driver. \\
R3 & \texttt{captive\_loosen} & Loosen the \{target\_name\} without releasing it from the panel and hold it in the loosened position. \\
R4 & \texttt{captive\_tighten} & Tighten the \{target\_name\} until it is seated, keep it captive, and withdraw the driver. \\
\bottomrule
\end{tabular}
\end{table*}

\begin{table*}[t]
\centering
\caption{Exact target IDs and instruction names. Each row is paired with every
listed template, giving 29 distinct task--target combinations.}
\label{tab:canonical_workcell_targets}
\footnotesize
\setlength{\tabcolsep}{4pt}
\renewcommand{\arraystretch}{1.04}
\begin{tabular}{@{}p{0.12\textwidth}p{0.28\textwidth}p{0.51\textwidth}@{}}
\toprule
Templates & Target ID & Exact \texttt{target\_name} \\
\midrule
M1--M5 & \texttt{Front\_quarter\_box} & front quarter box \\
M1--M5 & \texttt{Middle\_quarter\_box} & middle quarter box \\
M1--M5 & \texttt{Rear\_half\_box} & rear half box \\
R1 & \texttt{FDV1} & FDV1 cap \\
R2 & \texttt{SMA\_1\_1} & SMA cap at row 1, column 1 \\
R2 & \texttt{SMA\_1\_2} & SMA cap at row 1, column 2 \\
R2 & \texttt{SMA\_1\_3} & SMA cap at row 1, column 3 \\
R2 & \texttt{SMA\_2\_1} & SMA cap at row 2, column 1 \\
R2 & \texttt{SMA\_2\_2} & SMA cap at row 2, column 2 \\
R2 & \texttt{SMA\_2\_3} & SMA cap at row 2, column 3 \\
R2 & \texttt{SMA\_3\_1} & SMA cap at row 3, column 1 \\
R2 & \texttt{SMA\_3\_2} & SMA cap at row 3, column 2 \\
R2 & \texttt{SMA\_3\_3} & SMA cap at row 3, column 3 \\
R3, R4 & \texttt{captive\_1} & captive screw 1 \\
R3, R4 & \texttt{captive\_2} & captive screw 2 \\
\bottomrule
\end{tabular}
\end{table*}

\subsection{Initial states and terminal meaning}
Card transfer starts with the card displaced laterally from the opening;
insertion starts at the opening; extraction starts with the card inside the
selected box and the retention band around it. The other card tasks start
with the band released. Lid opening starts closed, whereas lid closing starts
open. Cap removal starts with the selected cap seated and threaded, its
adapter attached to the driver, and the stowage holder empty. Captive loosening
starts seated; captive tightening starts at the specified loosened thread pose.
The movable target must match its task ID: selecting a different SMA position
or captive screw rebuilds that target's articulated geometry while preserving
the source and target-frame bindings.

The geometric reset interface places these parts and exposes their named
frames. The environment supplies contact, elastic retention, attachment, and
joint measurements. In particular, the visible MELFI band pose is not a force
measurement, and a cap's adapter-relative pose is measured against the nominal
retained cap-to-adapter transform. The adapter frame follows the detachable
spindle after driver withdrawal. The canonical hardware protocol has a
30\,s horizon and a 1\,s terminal hold, so release, retention, docking, and
withdrawal are final-state requirements rather than labels inferred from a
scripted motion. Each episode binds the exact instruction, target identity,
asset manifest, and protocol into its specification hash.

\section{Workcell Hardware Details}
\label{app:hardware_details}

\begin{figure*}[t]
\centering
\includegraphics[width=\textwidth]{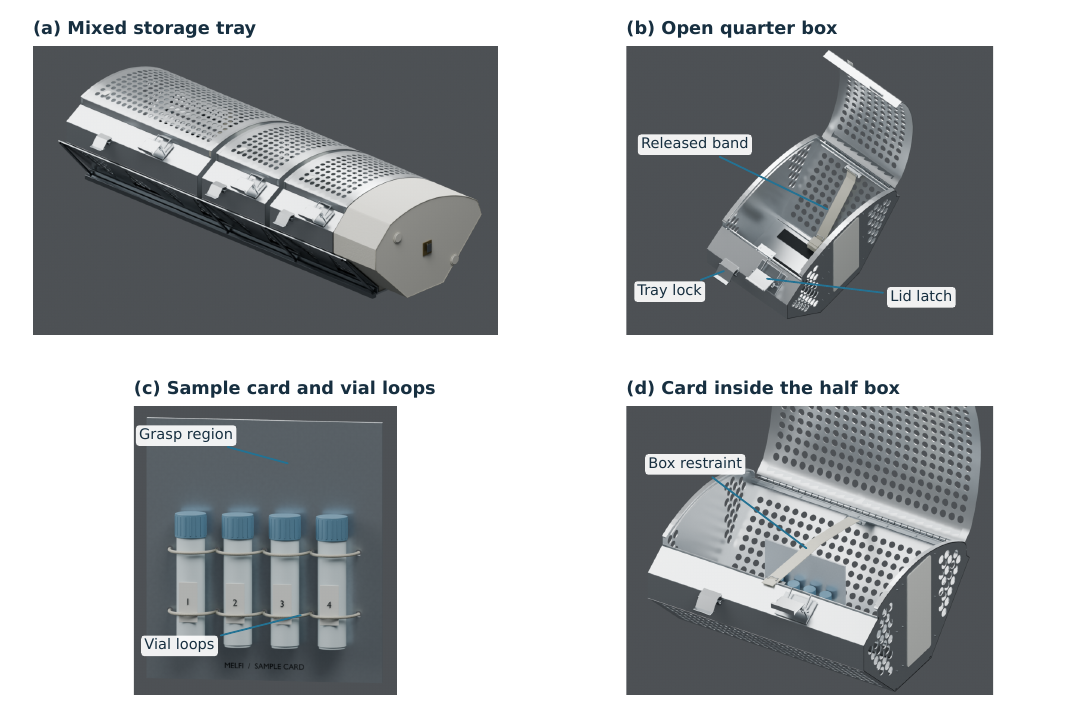}
\caption{MELFI storage hardware. (a) The mixed tray contains two quarter-size
modules and one half-size module, with separate lid and tray locks.
(b) An open quarter box exposes the hinge, curved perforated shell and
released card-restraint band. (c) The sample card carries four vials below
an unobstructed grasp region. (d) The half-box extraction reset places the
card inside with the band routed around it. Panels (b) and (d) show
task reset states. The historical mixed-box arrangement follows ESA and NASA
documentation; the card's $100\times110$\,mm planar dimensions follow
ESA's description~\cite{melfi_esa,melfi_nasa}.}
\label{fig:melfi_details}
\end{figure*}

\begin{figure*}[t]
\centering
\includegraphics[width=\textwidth]{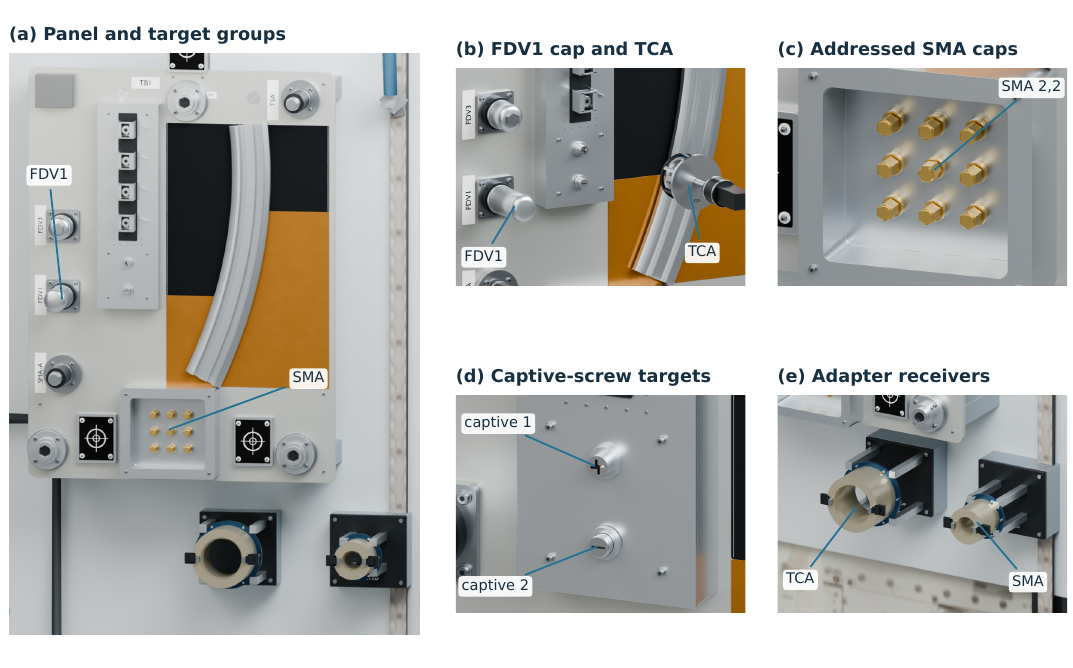}
\caption{RRM panel and tool interfaces in the accepted wall-mounted layout.
(a) The full board locates the FDV1 interface, SMA array, captive fasteners
and the two receivers below the panel. (b) The FDV1 cap and Tertiary Cap
Adapter are shown in the centered ready configuration.
(c) The $3\times3$ SMA array supplies addressed cap targets.
(d) The two captive-screw positions expose their drive interfaces.
(e) The larger TCA receiver and smaller SMA-adapter receiver contain
separate latch components. The views use the same final scene and part
poses; the arm and waiting tool are hidden where they obstruct the
hardware. NASA sources establish the reference panel and tool roles
\cite{rrm_thermal,rrm_nasa,rrm_2013}.}
\label{fig:rrm_details}
\end{figure*}

Figures~\ref{fig:melfi_details} and~\ref{fig:rrm_details} show close-up
renderings of the released models, exposing the openings, joints and
tool interfaces used in storage and servicing tasks.

\subsection{MELFI Boxes, Lids and Sample Card}
\textbf{Tray and box modules.}
The mixed tray follows the two-quarter-box and one-half-box arrangement
shown in ESA's 2002 MELFI description~\cite{melfi_esa}. The nominal
quarter-box envelope is $220\times130\times143$\,mm; the half-box envelope
is $220\times270\times143$\,mm~\cite{melfi_nasa}. These are maximum
envelopes: the curved, tapered box cross-section does not provide a
rectangular usable volume of the same size. The model preserves this
profile, the perforated surfaces, the tray rails and the insulation end
piece. NASA's 2015 interface document also records the subsequent phase-out of
quarter boxes~\cite{melfi_nasa}.

\textbf{Two different locking functions.}
Each box has a hinged lid, a lid latch and a separate tray lock.
The lid latch secures the cover; the tray lock belongs to the mounting
interface. Lid-opening and lid-closing tasks operate the lid latch and
hinge. The tray lock remains
part of the fixed installation. The authored open pose is $85^\circ$,
while the lid-opening task accepts an angle of at least $75^\circ$.
Closing requires the cover to return within $3^\circ$ of its closed
pose, with the lid latch engaged and the gripper withdrawn.

\textbf{Card, vials and restraint.}
ESA describes a $100\times110$\,mm aluminium vial card and elastic
restraints~\cite{melfi_esa}. The benchmark card is 2\,mm thick and carries
four vials, leaving a 25\,mm upper grasp region. These latter details
are authored dimensions. The vial loops belong to the card, whereas
the larger band belongs to the box and restrains the inserted card.
The band has one fixed end and one releasable end: the free end is
released for card passage and connected to restrain the card.
The released-band and around-card reset states make those roles visible
in Fig.~\ref{fig:melfi_details}. During insertion, the card must pass
through the opening with its vials and grasp interface clear of the
rim and lid; the final containment test covers the complete card
assembly. During extraction, the card must clear the opening by at
least 20\,mm. Transfer to the mouth instead ends at the named opening
frame.

\subsection{RRM Panel, Adapters and Receivers}
\textbf{Addressable interfaces.}
The TB1-inspired board contains three scored interface groups: the
FDV1 tertiary cap, nine SMA caps and two captive screws. The surrounding
caged fasteners, Marman interface and visual targets reproduce the
board's visible context~\cite{rrm_thermal}. The nominal
$400\times530\times6$\,mm board is an authored model scale. The SMA
targets are indexed by row and column. The default articulated asset
activates the central SMA cap and the first captive screw; selecting
another position rebuilds that target's movable geometry.

\textbf{Cap, adapter and driver are separate bodies.}
FDV1 removal uses a Multifunction Tool with a Tertiary Cap Adapter
(TCA). SMA removal uses a Safety Cap Tool and an SMA adapter
\cite{rrm_nasa,rrm_2013}. In the model, the driver housing, detachable
spindle, adapter retention parts and cap have distinct identities.
The tool engages the cap, rotates it off its thread, then withdraws
along the interface axis. The task requires at least 60\,mm of axial
withdrawal for FDV1 and 35\,mm for SMA. Those distances are benchmark
parameters. The adapter subsequently carries the cap into its matching
receiver.

\begingroup\clubpenalty=10000
\textbf{Receiver and fastener detail.}
Two receivers below the board accommodate the different adapter sizes.
Their guide cavities, flanges and latch parts explain the docking
geometry in Fig.~\ref{fig:rrm_details}(e). The cap task finishes with
the adapter secured and the driver detached and withdrawn. Captive
screws use a torque-set adapter and remain attached to the board.
Loosening requires at least 2.8 turns with 1.8--3\,mm axial travel;
tightening returns the screw to its seat before tool withdrawal.
\par\endgroup

\textbf{Installation and access.}
Four stand-offs attach the panel to a wall backplate. Its center is
983.45\,mm above the cabin floor. The robot mounting surface sits on a
28\,mm floor plate, centered 690\,mm in front of the panel plane. The
ready tool tip is 120\,mm in front of the panel center. These dimensions
describe the benchmark's cabin installation; the hardware reference is
the original exterior RRM experiment. Named tool and target frames
preserve the approach direction when changing the selected cap or
screw.

% Primary-source metadata verified on 2026-10-22.
% RoboCasa's proceedings record uses ``Household Tasks''; its arXiv/PDF title
% uses ``Everyday Tasks''. This entry follows the publisher's BibTeX record.

\end{document}